\documentclass[sigconf]{acmart}
\AtBeginDocument{%
  }

\copyrightyear{2025}
\acmYear{2025}
\setcopyright{acmlicensed}
\acmConference[CIKM '25] {Proceedings of the 34th ACM International Conference on Information and Knowledge Management}{ November 10--14, 2025}{Seoul, Republic of Korea.}
\acmBooktitle{Proceedings of the 34th ACM International Conference on Information and Knowledge Management (CIKM '25), November 10--14, 2025, Seoul, Republic of Korea}
\acmISBN{xxxxxx}
\acmDOI{xxxxxx}

\usepackage{hyperref}

\usepackage{balance}

\begin{document}

\title{MOVER: Multimodal Optimal Transport \\
with Volume-based Embedding Regularization}


\author{Haochen You}
\authornote{Corresponding author.}
\email{hy2854@columbia.edu}
\affiliation{
  \institution{
  Columbia University}
  \city{New York}
  \state{NY}
  \country{USA}
}

\author{Baojing Liu}
\email{liubj@hebic.edu.cn}
\affiliation{
  \institution{
  Hebei Institute of Communications}
  \city{Shijiazhuang}
  \state{Hebei}
  \country{China}
}



\begin{abstract}

Recent advances in multimodal learning have largely relied on pairwise contrastive objectives to align different modalities, such as text, video, and audio, in a shared embedding space. While effective in bi-modal setups, these approaches struggle to generalize across multiple modalities and often lack semantic structure in high-dimensional spaces. In this paper, we propose \textbf{MOVER}, a novel framework that combines optimal transport-based soft alignment with volume-based geometric regularization to build semantically aligned and structured multimodal representations. By integrating a transport-guided matching mechanism with a geometric volume minimization objective (GAVE), MOVER encourages consistent alignment across all modalities in a modality-agnostic manner. Experiments on text-video-audio retrieval tasks demonstrate that MOVER significantly outperforms prior state-of-the-art methods in both zero-shot and finetuned settings. Additional analysis shows improved generalization to unseen modality combinations and stronger structural consistency in the learned embedding space.

\end{abstract}




\maketitle

\section{Introduction}

With the rapid development of foundation models, multimodal learning has become a cornerstone for building general-purpose intelligent systems capable of understanding and generating content across diverse modalities such as vision, audio, and language \cite{zheng2024heterogeneous}. A central goal in this field is to construct a unified embedding space where semantically related data from different modalities are aligned \cite{wu2024comprehensive}, enabling tasks such as cross-modal retrieval, captioning, and audiovisual grounding \cite{zong2023self}. To this end, most existing approaches adopt contrastive learning frameworks that align pairs of modalities-typically using text as a common anchor-and optimize pairwise similarity in a latent space \cite{li2024multimodal,you2024application}.

\begin{figure}
\begin{center}
\includegraphics[width=\linewidth]{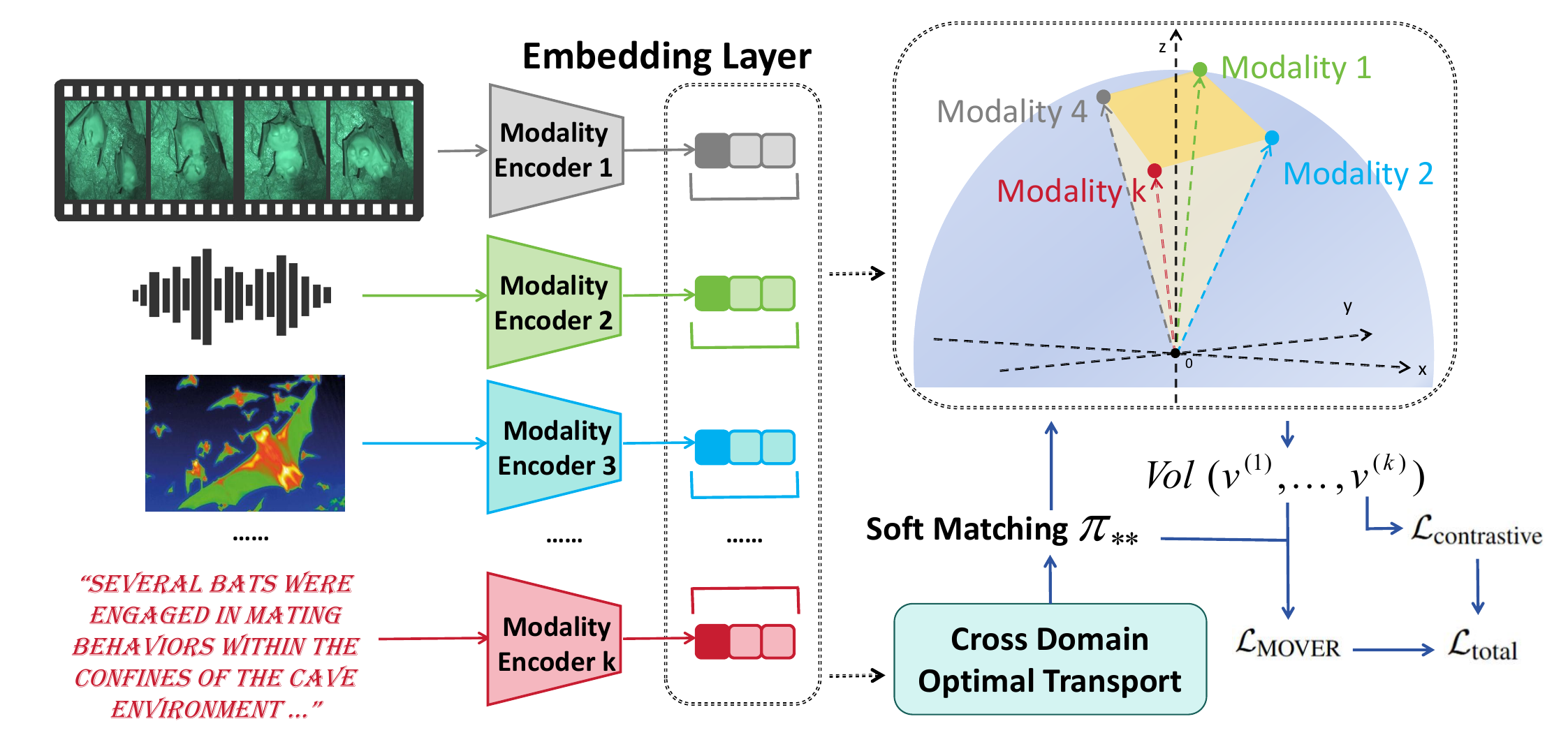}
\end{center}
\caption{Overview of the MOVER framework. Given inputs from $k$ modalities, modality-specific encoders produce unit-normalized embeddings, which are soft-matched via cross-modal optimal transport. GAVE module computes the geometric volume of matched groups as a measure of alignment. The final loss combines transport-based weighting and volume alignment to guide multimodal representation learning.}
\label{fig:illustration-total}
\end{figure}

While effective in bi-modal settings, pairwise strategies face clear limitations when extended to three or more modalities \cite{cicchetti2024gramian}. They often rely on fixed anchors or manual fusion, which may disrupt mutual alignment across modalities \cite{chen2025joint, zhang2024mm}. Moreover, similarity metrics like cosine similarity capture only local pairwise relations, ignoring global structure \cite{qian2025decalign}. This can lead to semantic inconsistency, modality dominance \cite{yu2024recent}, and entangled representations that hinder generalization and interpretability \cite{zhu2024enhancing}.


To address these challenges, recent efforts have explored geometric alignment mechanisms beyond simple pairwise matching \cite{wang2025otmkgrl, tjandrasuwita2025understanding}. For instance, volume-based objectives have been proposed to ensure embeddings from multiple modalities form coherent and semantically meaningful structures in latent space \cite{cicchetti2024gramian, ren2023importance}. While these approaches mark an important step forward, they still face two key limitations: their alignment mechanisms are typically based on hard matching or static pairing, lacking flexibility in modeling global semantic structures \cite{yuan2025survey}; moreover, it remains underexplored how to construct a principled and transferable alignment criterion in the absence of direct supervision for each modality pair \cite{zhang2024multimodal}.

In this paper, we propose \textbf{MOVER}, a new framework for multimodal representation learning that integrates optimal transport-based soft alignment with volume-based geometric regularization. By combining flexible many-to-many matching with compact volume preservation, MOVER promotes semantically consistent and structure-aware representations across modalities.

Our main contributions are as follows:
\begin{itemize}
    \item We introduce MOVER, a unified framework that leverages optimal transport and volume-based alignment (GAVE) for structured multimodal representation learning (Figure~\ref{fig:illustration-total}).
    \item We design an end-to-end contrastive objective that jointly enhances semantic discriminability and modality alignment, applicable in both zero-shot and finetuned regimes.
    \item We demonstrate state-of-the-art performance on multiple benchmarks, including text-video and text-audio retrieval, and validate the effectiveness of MOVER through ablations and visualization.
\end{itemize}

\section{Method}

\subsection{Multimodal Embedding Module (Encoders)}

We consider a $k$-modal representation learning setting, where the $m$-th modality corresponds to a data space $\mathcal{X}^{(m)}$ for $m = 1, \ldots, k$. Given a batch of multimodal samples, each sample consists of inputs from all $k$ modalities:
\begin{equation}
    \left(x_i^{(1)}, x_i^{(2)}, \ldots, x_i^{(k)}\right), \quad i=1, \ldots, B ,
\end{equation}
where $x_i^{(m)} \in \mathcal{X}^{(m)}$ denotes the raw input from the $m$-th modality for the $i$-th sample, and $B$ is the batch size.


For each modality \( m \in \{1,2,\ldots,k\} \), we define a modality-specific encoder \( E^{(m)}: \mathcal{X}^{(m)} \rightarrow \mathbb{R}^d \) that maps the input into a \( d \)-dimensional embedding vector. All outputs are $L_2$-normalized to ensure cross-modal comparability:
\begin{equation}
    \mathbf{v}_i^{(m)} = E^{(m)}\left(x_i^{(m)}\right) \big/ \left\| E^{(m)}\left(x_i^{(m)}\right) \right\|, \quad \forall i = 1, \ldots, B .
\end{equation}

As a result, each sample $i$ is represented by a set of $k$ unit-normalized embeddings:
\begin{equation}
    \left(\mathbf{v}_i^{(1)}, \mathbf{v}_i^{(2)}, \ldots, \mathbf{v}_i^{(k)}\right), \quad \mathbf{v}_i^{(m)} \in \mathbb{S}^{d-1},
\end{equation}
where $\mathbb{S}^{d-1}$ denotes the unit sphere in $\mathbb{R}^d$. This normalization ensures consistent scaling across modalities and stabilizes volume computation and optimal transport.

Each encoder $E^{(m)}$ adopts an architecture suited to its modality, and all encoders are jointly optimized using the unified alignment loss to learn semantically aligned representations in a shared space.

\subsection{Modality Matching Module}

After obtaining unit-normalized embeddings from all $k$ modalities, the next goal is to identify semantically aligned cross-modal sample groups, even when explicit alignment is not provided in the original data. We introduce an optimal transport (OT) mechanism to compute probabilistic matchings between modalities, producing soft-aligned candidate groups for subsequent geometric alignment.

Let $\mathcal{V}^{(m)}=\left\{\mathbf{v}_1^{(m)}, \ldots, \mathbf{v}_B^{(m)}\right\}$ denote unit-norm embeddings for the $m$-th modality in a batch, where $\mathbf{v}_i^{(m)} \in \mathbb{S}^{d-1}$. For each pair of modalities $p, q \in\{1, \ldots, k\}$, we define a cost matrix $C^{(p, q)} \in \mathbb{R}^{B \times B}$ based on the squared Euclidean distance between embeddings:
\begin{equation}
    C^{(p, q)}[i, j]=\left\|\mathbf{v}_i^{(p)}-\mathbf{v}_j^{(q)}\right\|^2.
\end{equation}

This cost matrix reflects the semantic discrepancy between samples across modalities. Given the cost matrix, we apply entropic regularized optimal transport (e.g., Sinkhorn algorithm) to solve for a soft matching matrix:
\begin{equation}
    \pi^{(p, q)}=\operatorname{Sinkhorn}\left(\mu^{(p)}, \mu^{(q)}, C^{(p, q)}, \varepsilon\right),
\end{equation}
where $\pi^{(p, q)} \in \mathbb{R}^{B \times B}$ denotes the soft matching weights from modality $p$ to $q$, $\mu^{(p)}$ and $\mu^{(q)}$ are uniform marginals over samples in $p$ and $q$, and $\varepsilon$ is an entropic regularization coefficient.


This process is repeated for all modality pairs, resulting in a set of soft matching matrices $\left\{\pi^{(p, q)}\right\}_{1 \leq p < q \leq k}$ defining probabilistic alignment between samples across modalities. Based on them, we generate candidate cross-modal groups using one of three strategies: \textbf{hard matching}, which selects the most probable match in each $\pi^{(p, q)}$ for every sample; \textbf{top-}$k'$\textbf{ matching}, which retains the top-$k'$ matches per modality pair; and \textbf{soft sampling}, which samples indices according to the matching distributions. The resulting combinations are passed to the geometric alignment module, which evaluates structural consistency via volume computation.




\subsection{Geometric Alignment Module}

In the framework of \textbf{GAVE} (Geometric Alignment via Volume Embedding), we measure cross-modal alignment by evaluating the geometric consistency of the modality embeddings in a shared latent space. Specifically, we compute the volume of the high-dimensional parallelotope spanned by the embeddings, which serves as an indicator of their directional alignment. A smaller volume implies that the vectors are more directionally aligned and thus semantically consistent across modalities.

We represent a matched sample group across $k$ modalities by stacking their unit-normalized embeddings into a matrix:
\begin{equation}
    V=\left[\mathbf{v}^{(1)} \mathbf{v}^{(2)} \cdots \mathbf{v}^{(k)}\right] \in \mathbb{R}^{d \times k}
\end{equation}
and compute the Gram matrix as $G=V^{\top} V \in \mathbb{R}^{k \times k}$.

The determinant of this Gram matrix corresponds to the squared volume of the parallelotope spanned by the $k$ vectors. Thus, the GAVE alignment measure is defined as:
\begin{equation}
    \operatorname{Vol}\left(\mathbf{v}^{(1)}, \ldots, \mathbf{v}^{(k)}\right)=\sqrt{\operatorname{det}\left(V^{\top} V\right)}=\sqrt{\operatorname{det}(G)} .
\end{equation}

Geometrically, this volume reflects how "spread out" the modality embeddings are in the shared space. A volume close to zero indicates that the vectors are nearly colinear or coplanar, implying strong alignment across modalities. In contrast, a larger volume suggests that the modalities diverge in direction, indicating weak alignment.

To ensure semantic consistency, we minimize the volume of each matched group, i.e., $\operatorname{Vol}\left(\mathbf{v}^{(1)}, \ldots, \mathbf{v}^{(k)}\right)$. This computation is differentiable and can be incorporated into network training as a loss term, enabling end-to-end optimization.


\subsection{MOVER Loss Construction}

In the MOVER framework, we integrate two complementary mechanisms: an optimal transport (OT) module that provides soft cross-modal matching, and a geometric volume regularization module (GAVE) that supervises the alignment quality. Together, these components define a unified training objective for learning a shared multimodal embedding space.

Let $\pi_{i_1, i_2, \ldots, i_k} \in[0,1]$ denote the soft matching probability produced by the OT module for a combination of samples $\left(i_1, i_2, \ldots, i_k\right)$ drawn from $k$ modalities. For each such combination, the geometric alignment is measured via the GAVE module as the volume of the parallelotope spanned by the corresponding embedding vectors.

We define $\mathcal{L}_{\text {MOVER}}$ as a weighted sum of volumes, using matching probabilities as weights:
\begin{equation}
    \mathcal{L}_{\text {MOVER}}=\sum_{i_1, \ldots, i_k} \pi_{i_1, \ldots, i_k} \cdot \operatorname{Vol}\left(\mathbf{v}_{i_1}^{(1)}, \ldots, \mathbf{v}_{i_k}^{(k)}\right).
\end{equation}

This loss encourages semantically matched sample groups (i.e., those with higher $\pi$) to exhibit tighter geometric alignment - i.e., lower volume and more colinear embeddings.

To enhance training stability, we add a contrastive loss. For each anchor sample from a chosen modality, we construct a positive group (via OT matching) and several negatives. Let $\operatorname{Vol}_{\text{pos}}$ and $\left\{\operatorname{Vol}_{\text{neg}}^{(j)}\right\}$ denote their respective volumes. The contrastive loss is defined as:

\begin{equation}
\mathcal{L}_i=-\log \frac{\exp \left(-\operatorname{Vol}_{\mathrm{pos}} / \tau\right)}{\sum_{j=1}^N \exp \left(-\operatorname{Vol}_{\mathrm{neg}}^{(j)} / \tau\right)},
\end{equation}
where $\tau$ is a temperature parameter controlling the sharpness.

We denote the mean contrastive loss over batch anchors as:
\begin{equation}
    \mathcal{L}_{\text {contrastive }}=\frac{1}{B} \sum_{i=1}^B \mathcal{L}_i .
\end{equation}

The total loss is a weighted combination of the two objectives:
\begin{equation}
    \mathcal{L}_{\text {total }}=\mathcal{L}_{\text {MOVER }}+\lambda \cdot \mathcal{L}_{\text {contrastive }} ,
\end{equation}
where $\lambda$ balances the contribution of the contrastive term. Depending on the application and empirical behavior, either one or both losses can be used during training.

Since all components in MOVER are differentiable, the model can be trained end-to-end via standard backpropagation. This allows the model to learn a unified embedding space in which semantically aligned samples across different modalities are geometrically aligned, leading to improved multimodal representation consistency and robustness.

\section{Experiments}

\subsection{Multimodal Semantic Retrieval}

To evaluate the semantic alignment capabilities of MOVER in capturing modality-consistent and semantically discriminative embeddings, we conduct cross-modal retrieval experiments on text-to-video (T2V / V2T) and text-to-audio (T2A / A2T) tasks across several benchmark datasets.

We evaluate on three datasets with diverse modality combinations and content: MSR-VTT \cite{xu2016msr}, AudioCaps \cite{kim2019audiocaps}, and VATEX \cite{wang2019vatex}. Modality-specific encoders are used to extract features-RoBERTa \cite{liu2019roberta} for text, ViT-B \cite{dosovitskiy2020image} for video frames, and BEATs \cite{chen2022beats} for audio. All embeddings are L2-normalized and projected into a unified embedding space. The MOVER model is trained using our proposed optimal transport-based soft matching mechanism, combined with the GAVE module’s geometric volume minimization objective, enabling structured alignment across modalities.

Table \ref{tab:zero-shot} reports the zero-shot performance of MOVER, where no task-specific finetuning is applied. MOVER consistently outperforms all baseline methods across all tasks, with Recall@1 improvements ranging from +2.8 to +4.5 percentage points. For instance, in the MSR-VTT T2V task, MOVER achieves a Recall@1 of 57.4\%, surpassing GRAM by +3.2\%. On the AudioCaps T2A task, the improvement reaches +4.5\%, the highest gain observed. These results demonstrate that MOVER constructs a more semantically aligned and generalizable embedding space, capable of capturing deep cross-modal relationships even without task-specific tuning.

Table \ref{tab:finetune} shows the performance after supervised finetuning. MOVER maintains strong superiority and further enlarges the performance gap. In the MSR-VTT T2V task, MOVER achieves 64.3\%, improving by +3.8\% over the best non-MOVER method, while in the A2T task on AudioCaps, the gain reaches +3.4\%. Although all methods perform strongly on VATEX, MOVER still achieves consistent gains, indicating its robustness in modeling stable multimodal representations across domains.

Overall, MOVER demonstrates systematic improvements under both zero-shot and finetuned settings. The results confirm that integrating optimal transport-based soft alignment with volume-based geometric regularization enables the construction of a unified, structured, and semantically meaningful multimodal latent space.

\begin{table}[htbp]
\centering
\caption{Zero-shot retrieval results (Recall@1, \%) on MSR-VTT, AudioCaps, and VATEX.}
\label{tab:zero-shot}
\small
\setlength{\tabcolsep}{6pt}
\renewcommand{\arraystretch}{0.9}
\begin{tabular}{l|cc|cc|cc}
\toprule
\textbf{Method} & \multicolumn{2}{c|}{\textbf{MSR-VTT}} & \multicolumn{2}{c|}{\textbf{AudioCaps}} & \multicolumn{2}{c}{\textbf{VATEX}} \\
 & T2V & V2T & T2A & A2T & T2V & V2T \\
\midrule
VAST \cite{chen2023vast}        & 47.2 & 45.9 & 38.4 & 37.0 & 74.2 & 75.1 \\
ImageBind \cite{girdhar2023imagebind}       & 51.0 & 52.3 & 39.8 & 38.9 & 75.0 & 76.3 \\
MM-Embed \cite{lin2024mm}        & 49.6 & 47.5 & 40.5 & 39.1 & 76.1 & 77.0 \\
VLM2Vec \cite{jiang2024vlm2vec}        & 50.4 & 48.2 & 41.3 & 40.2 & 77.5 & 77.8 \\
LanguageBind \cite{zhu2023languagebind}    & 51.1 & 48.7 & 47.1 & 41.0 & 78.2 & 78.6 \\
MuRAR \cite{zhu2024murar}                & 50.9 & 49.0 & 42.3 & 46.8 & 78.0 & 79.0 \\
Videoprism \cite{zhao2024videoprism}                & 52.0 & 49.5 & 43.8 & 42.7 & 78.8 & 80.2 \\
GRAM \cite{cicchetti2024gramian} & 54.2 & 51.6 & 45.2 & 43.9 & 81.1 & 79.5 \\
\midrule
\textbf{MOVER (Ours)} & 
\textbf{57.4} & 
\textbf{55.1} & 
\textbf{51.6} & 
\textbf{50.0} & 
\textbf{84.9} & 
\textbf{83.0} \\
\bottomrule
\end{tabular}
\end{table}

\begin{table}[htbp]
\centering
\caption{Finetuned retrieval results (Recall@1, \%).}
\label{tab:finetune}
\small
\setlength{\tabcolsep}{6pt}
\renewcommand{\arraystretch}{0.9}
\begin{tabular}{l|cc|cc|cc}
\toprule
\textbf{Method} & \multicolumn{2}{c|}{\textbf{MSR-VTT}} & \multicolumn{2}{c|}{\textbf{AudioCaps}} & \multicolumn{2}{c}{\textbf{VATEX}} \\
 & T2V & V2T & T2A & A2T & T2V & V2T \\
\midrule
VAST \cite{chen2023vast}        & 55.2 & 54.1 & 43.8 & 42.7 & 83.4 & 81.7 \\
ImageBind \cite{girdhar2023imagebind}       & 57.0 & 57.7 & 43.3 & 42.6 & 84.2 & 82.5 \\
MM-Embed \cite{lin2024mm}        & 56.8 & 55.1 & 44.1 & 43.0 & 84.5 & 83.0 \\
VLM2Vec \cite{jiang2024vlm2vec}        & 57.5 & 55.9 & 45.0 & 44.2 & 85.1 & 83.5 \\
LanguageBind \cite{zhu2023languagebind}    & 58.3 & 56.2 & 46.4 & 45.1 & 87.3 & 83.8 \\
MuRAR \cite{zhu2024murar}                & 58.1 & 58.3 & 45.5 & 44.6 & 85.3 & 85.2 \\
Videoprism \cite{zhao2024videoprism}                & 60.5 & 57.8 & 46.9 & 45.9 & 86.0 & 84.5 \\
GRAM \cite{cicchetti2024gramian} & 58.6 & 57.4 & 49.2 & 48.0 & 86.3 & 84.6 \\
\midrule
\textbf{MOVER (Ours)} & 
\textbf{64.3} & 
\textbf{61.1} & 
\textbf{53.2} & 
\textbf{51.4} & 
\textbf{89.1} & 
\textbf{87.1} \\
\bottomrule
\end{tabular}
\end{table}

\subsection{Embedding Space Visualization Analysis}

To assess the effectiveness of MOVER in cross-modal semantic alignment, we conduct a t-SNE-based visualization to analyze the high-dimensional embedding space learned by different models.


We evaluate on three datasets with explicit class labels and multiple modalities-VGGSound \cite{chen2020vggsound}, AudioSet \cite{gemmeke2017audio}, and Kinetics-700 \cite{carreira2019short}-by selecting the top three categories from each and extracting corresponding text (as semantic anchors), video, and audio.


\begin{figure*}[t]
\begin{center}
\includegraphics[width=\linewidth, height=0.17\linewidth]{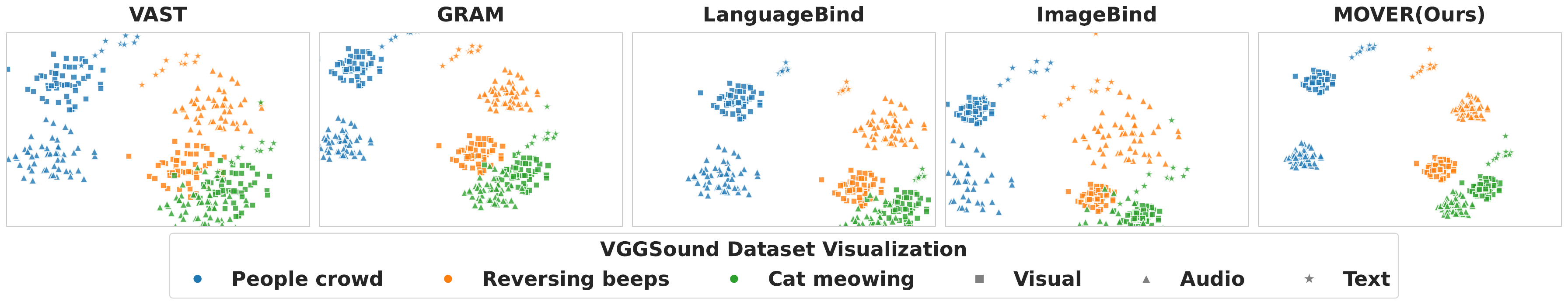} \\
\end{center}
\caption{t-SNE visualizations for VGGSound.}
\label{fig:tsne-double}
\end{figure*}

Figure~\ref{fig:tsne-double} illustrates the embedding distributions for VGGSound. VAST shows scattered clusters, with audio drifting away from text.
GRAM improves cohesion but exhibits overlapping class boundaries. LanguageBind clusters well around text but fails to align audio and video fully. ImageBind handles the visual modality effectively but lacks clear semantic separation across modalities.

In contrast, MOVER produces the most structured embeddings across all datasets: audio, video, and text are tightly clustered per category with clear inter-class separation. In AudioSet, audio and text nearly overlap at the category centers; in Kinetics-700, modality convergence patterns reflect complex action semantics. These results highlight MOVER’s ability to capture both inter-modal consistency and semantic differentiation.

\subsection{Cross-Modality Generalization}

To assess the generalizability of MOVER to unseen modality combinations, we design a cross-modality generalization experiment. Models are trained solely on text-video (T-V) pairs, with no access to audio during training. At test time, we evaluate their zero-shot performance on text-audio retrieval tasks (T2A and A2T), requiring the model to infer text-audio alignment from T-V training alone.

We evaluate on four T-A datasets-AudioCaps, VATEX, Clotho \cite{Drossos_2020_icassp}, and SoundDescs \cite{Koepke_2021_arxiv}-and compare MOVER to baselines. An oracle model trained with both T-V and T-A data serves as an upper-bound reference. As shown in Table~\ref{tab:cross-modal}, MOVER consistently outperforms all baselines in zero-shot T-A retrieval, despite never observing audio-text pairs during training. It improves Recall@1 over GRAM by an average of 3.9 points across datasets and directions, demonstrating its ability to generalize across unseen modality pairs.





\begin{table}[htbp]
\centering
\caption{Cross-modality generalization results (Recall@1, \%) on four T-A retrieval datasets. All models are trained only on T-V pairs and evaluated on T-A tasks.}
\label{tab:cross-modal}
\small
\setlength{\tabcolsep}{3.5pt}
\renewcommand{\arraystretch}{0.9}
\begin{tabular}{l|cc|cc|cc|cc}
\toprule
\textbf{Method} & \multicolumn{2}{c|}{AudioCaps} & \multicolumn{2}{c|}{VATEX} & \multicolumn{2}{c|}{Clotho} & \multicolumn{2}{c}{SoundDescs} \\
 & T2A & A2T & T2A & A2T & T2A & A2T & T2A & A2T \\
\midrule
VAST \cite{chen2023vast}          & 37.2 & 35.9 & 71.0 & 69.8 & 40.5 & 38.7 & 42.2 & 40.9 \\
LanguageBind \cite{zhu2023languagebind} & 39.1 & 37.6 & 72.3 & 71.5 & 42.0 & 40.1 & 43.9 & 42.4 \\
ImageBind \cite{girdhar2023imagebind}   & 38.4 & 36.8 & 73.0 & 72.1 & 42.5 & 41.3 & 44.1 & 42.6 \\
GRAM \cite{cicchetti2024gramian}        & 42.5 & 41.2 & 75.2 & 74.0 & 46.3 & 44.5 & 47.0 & 45.6 \\
\textbf{MOVER (Ours)}                  & 
\textbf{46.8} & 
\textbf{45.3} & 
\textbf{78.6} & 
\textbf{77.4} & 
\textbf{50.2} & 
\textbf{48.7} & 
\textbf{51.0} & 
\textbf{49.3} \\
\midrule
Oracle (T-V + T-A)                     & 51.6 & 50.3 & 81.1 & 79.5 & 53.4 & 51.5 & 54.0 & 52.6 \\
\bottomrule
\end{tabular}
\end{table}

While the oracle model still achieves the highest scores, the small gap (typically 2-4 points) suggests that MOVER effectively captures transferable semantic structures. These results validate that our optimal transport-based matching and GAVE alignment together support robust and modality-agnostic representation learning.



\subsection{Ablation Study}

\begin{table}[htbp]
\centering
\caption{Ablation study on MSR-VTT, AudioCaps, and VATEX (Recall@1, \%) evaluating OT and GAVE modules.}
\label{tab:ablation}
\footnotesize
\setlength{\tabcolsep}{5pt}
\renewcommand{\arraystretch}{0.85}
\begin{tabular}{l|ccc}
\toprule
\textbf{Variant} & MSR-VTT (T2V) & AudioCaps (T2A) & VATEX (T2V) \\
\midrule
MOVER (full model)           & \textbf{57.4} & \textbf{51.6} & \textbf{84.9} \\
w/o OT (cosine only)   & 54.1 \textcolor{blue}{(--3.3)} & 47.2 \textcolor{blue}{(--4.4)} & 81.2 \textcolor{blue}{(--3.7)} \\
w/o GAVE (no volume)   & 53.7 \textcolor{blue}{(--3.7)} & 46.8 \textcolor{blue}{(--4.8)} & 80.6 \textcolor{blue}{(--4.3)} \\
w/o both               & 51.9 \textcolor{blue}{(--5.5)} & 44.5 \textcolor{blue}{(--7.1)} & 78.5 \textcolor{blue}{(--6.4)} \\
\bottomrule
\end{tabular}
\end{table}


Table \ref{tab:ablation} shows the results of our ablation study on MSR-VTT, AudioCaps, and VATEX. Removing either the OT soft matching or the GAVE volume alignment leads to a clear drop in performance across all datasets. These results confirm that both modules are critical and complementary: OT encourages flexible semantic alignment across modalities, while GAVE regularizes the embedding space by enforcing geometric consistency. Their combination is essential for building a robust and generalizable multimodal representation.

\section{Conclusion}

We propose \textbf{MOVER}, a framework that combines optimal transport-based soft alignment with volume-based geometric regularization for multimodal representation learning. MOVER captures global semantic structure beyond pairwise similarity and consistently outperforms prior methods on zero-shot and finetuned retrieval tasks. It also generalizes well to unseen modality combinations and maintains robust alignment in high-dimensional embedding spaces. These results highlight the promise of integrating transport-based matching with geometric modeling for unified multimodal learning.

\section*{GenAI Usage Disclosure}
Generative AI tools, specifically ChatGPT, were used during this research for minor language polishing and code debugging support. These tools did not contribute to the design, execution, or analysis of the research study, nor to the generation of any original scientific content. The authors retained full control and responsibility over all aspects of the research and manuscript preparation.

\bibliographystyle{ACM-Reference-Format}
\balance
\bibliography{sample-base}

\end{document}